# Hallucination is the last thing you need

**Travers Smith**

**Shawn Curran**
**Travers Smith**
shawn.curran@traverssmith.com

**Oliver Bethell**
**Travers Smith**
oliver.bethell@traverssmith.com

**Sam Lansley**
**Travers Smith**
sam.lansley@traverssmith.com

## Abstract

The legal profession necessitates a multidimensional approach that involves synthesizing an in-depth comprehension of a legal issue with insightful commentary based on personal experience, combined with a comprehensive understanding of pertinent legislation, regulation, and case law, in order to deliver an informed legal solution. The present offering with generative AI presents major obstacles in replicating this, as current models struggle to integrate and navigate such a complex interplay of understanding, experience, and fact-checking procedures. It is noteworthy that where generative AI outputs understanding and experience, which reflect the aggregate of various subjective views on similar topics, this often deflects the model's attention from the crucial legal facts, thereby resulting in hallucination. Hence, this paper delves into the feasibility of three independent LLMs, each focused on understanding, experience, and facts, synthesising as one single ensemble model to effectively counteract the current challenges posed by the existing monolithic generative AI models. We introduce an idea of mutli-length tokenisation to protect key information assets like common law judgements, and finally we interrogate the most advanced publicly available models for legal hallucination, with some interesting results.

## 1 Introduction

There are various mechanisms under way to try and resolve hallucination errors within LLMs, including enhancing the models understanding of facts [1] Meng and Bau et al, having models call out under a retrieval mechanism for facts [2] Izacard and Grave or even in-industry Legal Tech vendors using search and comparison algorithms to fact check generative output [3] Lexis+ AI.

Hallucinations are already causing issues within the legal industry, with a lawyer in the US awaiting a sanctions hearing for [4] presenting hallucinated case law in front of the courts. One court has responded with a [5] standing order to ensure lawyers inform them where generative AI has been used in their submissions.

Use of generative AI for legal research potentially impacts the integrity of common law, as we discuss in this [6] article. In our view, common law contamination with subtle, non-obvious errors poses the biggest threat to the credibility of the legal industry in using generative AI for legal research.

Therefore, we expect a great deal of investment to now go into fixing the hallucination problem, backed up by [7] OpenAI themselves recently and presumably a follow up from the hallucinated case law issue. ChatGPT has served as a great demonstration to the industry of the sophistication of generative AI, and our view is that firms should leverage our open source educational and experimental chatbot [8] YCNBot to ensure your business keeps up to date and informed on this technology until it's ready for primetime.

As previously discussed, the resolution of hallucinations in legal settings requires multiple approaches. This paper proposes a theoretical approach that we believe holds promise, and we intend to explore this approach in-depth as our data labelling processes reach their critical mass.

We wish to clarify that while the authors of this paper possess some formal legal training, we are not qualified lawyers. The opinions and perspectives presented in this paper draw from legal data, given this is our domain, however we approach this on a computational basis to assess the limitations of LLMs. We do not provide legal advice or judgement. Our firm has a leading Technology law practice that we collaborate closely with. This practice is available to assist any organisation with legal matters tied to their AI strategy. For further details, kindly contact james.longster@traverssmith.com or louisa.chambers@traverssmith.com.

## 2 Common law contamination

As previously indicated, a spurious instance of case law came to the attention of the court and was presented before a perplexed judge. This was readily apparent to the court due to the absence of the case name. However, what if generative AI is being employed more expansively through the legal sector, where subtle, non-obvious errors are surreptitiously inserted into motions, subsequently adopted into judgments, and conventional law evolves such that the facts of a judgment, repeatedly cited, gradually transform, and become altered with slight, subtle changes that ultimately substantively affect the initial judgment over time. Never before has the legal profession possessed a product that makes a statistical prediction of the subsequent word in a judgment, presumptuously overruling the court's actual words with 'hallucination'. In this paper, we suggest a method for curbing hallucinations in the legal domain around architectural design of models, and then examine the apparition and hallucination issues, that we contend support the necessity of exercising extreme caution in using LLMs for legal research.

Our understanding of the internal structure of OpenAI's architecture is limited. We acknowledge the presence of instructional, completion, question and answer, classification, and summarisation components which work together to tackle generative AI issues. Nevertheless, we believe that within the larger LLMs, which have been trained on a vast amount of publicly available data, they exhibit characteristics of being a "jack of all trades, hallucination in some" whereby they are monolithic and exhibit intricate interplays between various features such as case law, legislation, and multiple case summaries written in diverse styles resulting in them computing unclear and overlapping output across each. We recognize that several different approaches can be used to address these challenges, and we propose two alternative solutions - multi-length tokenization and a vertically aligned ensemble models.

## 2.1 Multi-length tokenisation

An essential component of constructing LLMs is the tokenization process [9] [Mielke, Alyafeai et al] which involves training a model to predict the next most probable word based on the input. Our concern is specifically with its application in the common law domain.

In this context, a judgment should be represented verbatim as a factual database, rather than relying on a probabilistic guess the next word approach. This becomes particularly important when considering quotes from case law, for example, where a judge may make a statement such as: *The tenant claimed negligence under Tort accusing the landlord of not resolving a defect in the property for a period of x years. It is my view that where notice has been provided to the landlord from the tenant and a reasonable period has passed, the exposure of the landlord to negligence increases, as discussed in X v Y [2012]. **"an immediate attempt to resolve a defect that has a natural and unfortunate delay cannot be considered equal to prolonged ignorance when notice of the defect has been given"**.*

It is abundantly clear that the text within a judgment is of paramount importance. If we were to request the model to ingest the facts of a case, consider the arguments on both sides, and provide an opinion based on that reasoning process, some level of creativity or hallucination may be acceptable. However, the information that binds together every judicial decision, the quote made bold in the previous paragraph, is of the utmost significance, and we argue should be given "structured data" status. As such, we advocate this aspect of common law data needs to be excluded from the probabilistic guesswork applied in LLMs.

To provide a brief illustration of how tokenization operates at the word level, consider the example of training the model that the phrase "When for instance one" is often followed with the word "condition" However, this probabilistic prediction method is not well-suited to handling quotes within legal judgments. For instance, as the LLM progresses into the next token which might, for example, be "is" then "onerous" if we tokenize the words individually within the quote, we run the risk of incorrectly representing the quote. Alternatively, if we tokenize the entire quote, the predictive output could be completely off-topic based on predictions derived from preceding text. This creates a significant difficulty when tokenizing complex legal texts that cannot be easily broken down into individual words or phrases, yet which must be treated as integral units in their own right.

| | |
|---|---|
| When | for |
| When for | instance |
| When for instance | one |
| When for instance one | condition |
| When for instance one condition | in |
| When for instance one condition in | a |
| When for instance one condition in a | set |
| When for instance one condition in a set | is |
| When for instance one condition in a set is | particularly |
| When for instance one condition in a set is particularly | onerous |
| When for instance one condition in a set is particularly onerous ... | "Some clauses which I have seen would need to be printed in red ink on the face of the document with a red hand pointing to it before the notice could be held to be sufficient." |

Although we will be researching this approach, we recognise the challenges this will face for LLM's. Some of which are (1) the context leading up to the quote will often always be different, relating to the facts of the case on that judgement, (2) the entire quote might not always be needed, and if subsets of common quotes are also common, the output might not fully support being bound to the facts of the current case.

## 2.2 Ensemble models

Another theoretical approach we are researching is an ensemble model, which is a collection of models trained on different data working together to improve the quality of the models for legal research.

Using the example provided below: when constructing a persuasive argument, one typically begins with the blue phase – a description of the underlying problem. This may involve examining the events of a criminal investigation, assessing the organisational structure of a large company as part of a merger, or understanding details of an issue from a client grappling with complex regulatory issue. Such an analysis constitutes the "understanding" phase of argumentation, as it seeks to establish the fundamental issue at hand. Moving to the yellow phase, one is able to draw upon insights gleaned from experience and comprehension to offer thoughtful commentary. Judges may use previous rulings to inform their decisions in such cases, while lawyers might link a client's problem to relevant legislation or case law when offering advice. Finally, the green phase involves the direct quotation from a relevant judgement or legislation to support and add weight to the argument. Taken together, these phases form a robust foundation upon which an effective argument can be built.

Stiletto Visual Programmes (SVP) ordered 47 photographic transparencies from Interfoto Picture Library (IPL). On the delivery note was a clause stating that transparencies should be returned within 14 days of delivery. If they were not so returned, a holding fee of £5 per transparency per day would be charged. SVP returned the transparencies four weeks later and received a bill for over £3,700. SVP refused to pay. SVP contended they had never dealt with IPL before, were unaware of their standard conditions and they had not been sent a copy of their conditions prior to their having returned the transparencies. Now as regards each of the plaintiffs, if at the time when SVP accepted the ticket, SVP, either by actual examination of it, or by reason of previous experience, or from any other cause, was aware of the terms or purport or effect of the endorsed conditions, it can hardly be doubted that SVP became bound by them. I think also that SVP would be equally bound if he was aware or had good reason to believe that there were upon the ticket statements intended to affect the relative rights of himself and the company, but intentionally or negligently abstained from ascertaining whether there were any such, or from making themselves acquainted with their purport. But I do not think that in the absence of any such knowledge or information, or good reason for belief, SVP was under any obligation to examine the ticket with the view of ascertaining whether there were any such statements or conditions upon it. More recently the question has been discussed whether it is enough to look at a set of printed conditions as a whole. When for instance one condition in a set is particularly onerous does something special need to be done to draw customers' attention to that particular condition? In J. Spurling Ltd. v. Bradshaw [1956] 1 W.L.R.461 Lord Justice Denning stated that "Some clauses which I have seen would need to be printed in red ink on the face of the document with a red hand pointing to it before the notice could be held to be sufficient".

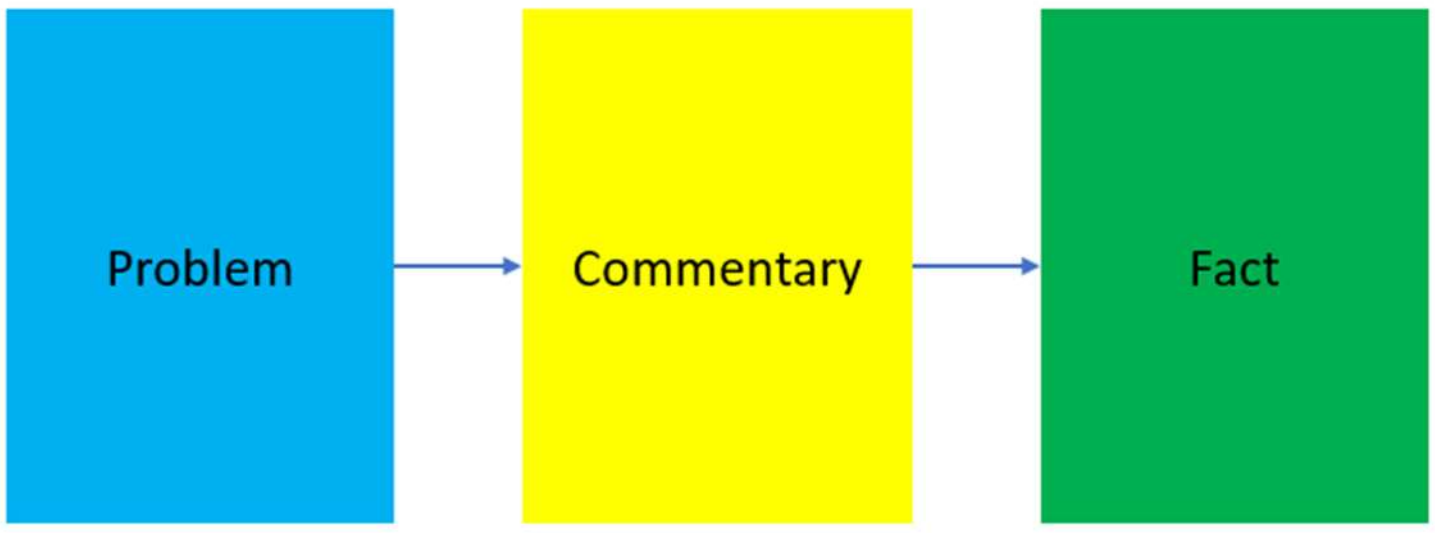

**Problem Model**

It is important to note that the "problem" model ought to be trained on the early section of the judgement, specifically the summary of the case. The inclusion of a detailed and comprehensive account of the claims and defences presented throughout the trial would undoubtedly strengthen the model has it access to that data also. It is important to bear in mind that the primary objective is to teach the model to appropriately learn about "problems". This relationship can be likened to a question-and-answer pair albeit on a much larger scale, with the question representing the problem as input and the answer representing the commentary as output.

**Commentary Model**

The commentary model essentially functions in a way similar to human experience. While some may argue for a division of this model, we subscribe to the thinking in [10] Dasgupta, Lampinen et al, which highlights the distinction between a fact database (e.g. System 1) and a reasoning engine (e.g. System 2). Our brains automatically send a given problem to either of these systems to find a viable solution. To illustrate this point, we consider the example of "What is the capital of France?" where the answer would be a problem-based fact extraction, namely Paris. Converse to that, the question "What is the capital of Australia?" has historically been rather tricky, with cities such as Sydney, Melbourne, or Perth being preferred before the correct answer, Canberra, became commonly known. In instances such as these, our logical "System 2" brain comes into play as we seek to deduce the answer based on properties such as the largest city, the most popular, and similar patterns we find in other countries such as Paris, London, or Madrid.

**Fact Model**

In the given context, the fact model pertains to the verbatim quote used within the judicial process to augment an argument, be it as part of counsel making arguments or as part of a judgement. Its primary function is to lend additional weight to an argument, lending credibility and persuasive power to the overall framework of reasoning or decision-making at hand.

If the commentary model is system 2, then the problem and fact models are system 1.

**Linking them together**

The process of determining when to pass the entire problem text to the commentary model in the problem model can be complex. One possible solution is to employ a process of breaking down judgements into three types - Problem, Commentary, and Fact - thereby creating an <EOP> End of Problem entity that triggers a <SOC> Start of Commentary token. This would in turn signal a controlling algorithm in the ensemble to pass the problem text to the commentary model. The algorithm would understand that it is currently working on an entity of type problem and building an entity of type commentary until it reaches an <EOC> End of Commentary tag. At this point, it would move into the fact category. However, it should be noted that not all <EOC>s would be followed by a <SOF> Start of Fact token, and the model may need to alternate between problem, commentary, and fact segments in a non-linear fashion.

Although this approach can be complicated due to humans' natural fluidity in moving through such segments, implementing a framework where P>C>F represents three known pairs for each argument in a judgement could make it easier to break open the monolithic single LLM and have domain-defined vertical models working together. It is worth noting that the idea of having a 'System 1' and 'System 2' already suggests that our brains group different sets of neurons for different tasks, underscoring the potential for segregated models in the context of these entities.

An obvious "hand-over" pattern to the fact model would be:

Fact Hand-Over Patterns

| |
|---|
| the tribunal relied on the well-known dictum of Denning LJ in Spurling (J) Ltd. v Bradshaw [1956] 1 WLR 461, at 466, that: |
| As Lord Denning said in J Spurling Ltd v Bradshaw [1956] 1 WLR 461 at 466: |
| graphically described by Denning LJ in J Spurling Ltd -v- Bradshaw [1956] 1 WLR 461: |
| observed in J. Spurling Limited v. Bradshaw [1956] 1 WLR 461 at 466: |
| In an obiter dictum in J. Spurling Ltd. v. Bradshaw [1956] 1 W.L.R.461 at page 466 (cited in Chitty on Contracts 25th Ed. Vol. 1 at page 408) Lord Justice Denning stated that |

A key challenge here is determining whether the commentary model fabricates case law in the "fact handover pattern." Is the reference to the case and section of the verdict employed to substantiate the fact (the quoted evidence bolstering the argument), or is it employed as a template and model feature for guessing the subsequent token, which may be an entire fact from the fact model. Potentially, both paths converge towards the same result, whether through rule-based and search strategies or probabilistic ones, and the hurdle lies in guaranteeing that the commentary model refrains from generating fictitious information. If an upstream model hallucinates, a downstream model hallucinates, though with this approach, it hallucinates by presenting chunks of text, not individual tokens. This "hallucination" wouldn't make sense but would represent fact verbatim.

A conceivable strategy to overcome this challenge could involve refraining from incorporating the entire paragraph, or even the complete problem text into the fact model. Instead, the focus may shift towards solely considering the final sentence, as it functions akin to a Key/Value pair while linking the commentary of "J. Spurling Ltd v Bradshaw [1956] WLR 461 at 466" with the fact of "big red hand […] red ink […]". However, binding ensemble models through key value pairs seems to be a sticking plaster.

Taking some examples from the commentary text preceding the case law quotations that are similar, there is clearly a lot of semantically similar intent:

| |
|---|
| whether it is enough to look at a set of printed conditions as a whole |
| particularly onerous or unusual conditions are brought fairly and reasonably to the attention of the other party |
| He contends that the IAC is so onerous and exceptional in its provisions that it cannot be incorporated by a mere reference because specific and particular notice was required |
| expressed the opinion that he should consider the plaintiff had given sufficient notice of this condition |

The features here are "conditions", "onerous", "unusual", "notice", "sufficient" which when attended to, almost perfectly lines up for the more "onerous" and "unusual" a "condition" "sufficient" "notice" should be provided.

Therefore, semantically similar commentary, prepared through tokenisation of LLM's but when fully built as a sentence, is taken back into a classification model or key/value pair to predict the next section of text, the case law citation, followed by the fact model which provides the entire verbatim quote, with the fact model trained as the next sequence based on some larger sections of preceding text – e.g. the semantic commentary and the citation text.

As it is evident, the process of linking these intricate models would present a formidable challenge. Undeniably, our primary objective is to safeguard the quotations utilised in judgments from any obscure or ambiguous hallucinations. Consequently, the proposed solution places significant emphasis on addressing this crucial issue.

**Judgement as Fact Issue**

Undoubtedly, there exists a salient point that must be taken into consideration, namely that a judgment may contain certain elements that cannot be cited, and consequently, a subsection of that judgment may never again be referenced in another judgment. We must acknowledge that while an entire judgment may be construed as a "fact", training the fact model with an entire judgement may detract from the importance of adequately and thoroughly respecting the more frequently cited quotation text. Additionally, we want to ensure that irrelevant text in a judgment, which has never been referenced in a subsequent judgment, does not negatively impact the fact model. The possibility arises that the model could encounter challenges if a new notable judgment were to emerge, however has not been cited or referenced in subsequent judgments yet. One potential solution to this issue involves the courts proactively enhancing their data labelling techniques, thereby simplifying the way we obtain relevant information for the model's consumption to support ongoing finetuning of the ensemble models.

# 3 Hallucination Evaluation

With the aim of evaluating the performance of GPT models, we will leverage OpenAI as a testing ground. Specifically, we will focus on assessing the model's accuracy in reproducing factual content from legal judgments while also exploring its propensity for generating false information ("hallucinations"). Through this process, we hope to gain a comprehensive understanding of the capabilities and limitations of the GPT models in their ability to accurately quote verbatim fact from judgments.

Using the Complete mode of Playground with the text-davinci-003 model we get the following results:

> Spurling v. Bradshaw [1956] 3 All E.R. 121 was a case in which the defendant sent to the plaintiff, who carried on a business as a warehouse-man, eight barrels of orange juice which due to his negligence were badly damaged. The plaintiff sued for his charges and in answer to the defendant's counterclaim for damages pleaded a clause in a "landing account" which he had delivered to the plaintiff which exempted him from liability for any loss howsoever, whensoever, and wheresoever even if occasioned by his servants or agents. In the course of his judgment, Denning L.J. expressed the opinion that he should consider the plaintiff had given sufficient notice of this condition to the defendant and only if he had would the claim be enforceable. In holding that this had been done he observed that the more unreasonable the clause, the greater the notice which must be given of it adding:-
>
> "The justness of it is often the best proof of whether it has been sufficiently brought to the other's notice."

> [24] Most disputes which arise relate to the third test and the issue of reasonable sufficiency of notice, namely whether the party tendering the document has done all that was reasonably sufficient to give the other notice of the conditions. Whether they did so or not is a question of fact (Chitty 13-014). It is well established that where the party tendering the document knows it contains an onerous or unusual term it must show that it has been brought fairly and reasonably to the other's attention (Chitty 13-015). As Lord Denning said in J Spurling Ltd v Bradshaw [1956] 1 WLR 461 at 466:
>
> `The law is quite clear that, if one party to a contract introduces a printed form containing printedterms, the other is not bound by these terms unless he has had a reasonable opportunity of knowing the contents of the form'.

> The references there are to the exchange of letters on 19 and 20 October 1998. The tribunal's finding was that that exchange constituted the making of Mr Sweeney's contract of employment, which then started on 26 October. They held that, by that exchange, Peninsula incorporated into the contract the commission scheme rules which had been referred to by Mr Done, but which it had not first provided to Mr Sweeney and which he had not seen. They held next that, in the light of the facts to which we have referred (the time lag in the payment of commission earned on most sales), the terms of section B(1), (2) and (3) were unduly onerous, and that Peninsula could only rely on them if it could show that it had first fairly brought them to Mr Sweeney's notice, which the tribunal found it had not. In reaching this conclusion, the tribunal relied on the well-known dictum of Denning LJ in Spurling (J) Ltd. v Bradshaw [1956] 1 WLR 461, at 466, that:
>
> "No man is bound to abide by any term or condition of which he was unaware, and had no reasonable opportunity of knowing before he made the contract."

It is evident that the GPT models rely heavily on the input sequence and textual context to anticipate the subsequent section rather than producing an exact replica of the forthcoming statement, such as the red ink quote. Nonetheless, the output sentiment demonstrates similarities in regard to phrases such as "*sufficiently brought to the others' notice*", "*reasonable opportunity of knowing the contents*", and "*term or condition of which he was unaware."* These phrases arguably convey semantic equivalences to the original quote of "*Some clauses I have seen would need to be printed in red ink on the face of the document with a red hand pointing to it before the notice could be held to be sufficient.*" This observation indicates that older models of GPT might have assigned more weight to their own interpretation of the facts instead of the precise wording used in the judgment. There is also the possibility that Temperature has been a contributing factor, making previous variants more imaginative and less definitive.

In order to advance the evaluation process, the evaluation is transferred to GPT-4 within the Chat Mode of Playground, where the temperature is adjusted to 0 along with a predetermined prompt:

---

*Finish this sequence with the quote that follows this paragraph:*

*"Consideration of the effect of an exclusive jurisdiction clause is central to the assessment as to whether or not it is an unusual or onerous term I think that the number of cases in which exclusive jurisdiction clauses come before this court is testament to the fact that they are not unusual. Are they, however, onerous? Paragraph 1 of Schedule 4 to the Civil Jurisdiction and Judgments Act 1982 provides "*

---

The purpose of this eval is to test whether the model, with enough preceding context, knows to output the follow-on quote verbatim.

The eval should test the following:

- Quote Check
  - Verbatim match with the source judgement.
  - Close to verbatim match including accompanying text.
  - Non-verbatim match with similar semantic intent.
  - Non-verbatim match with unrelated intent.
- Hallucination Check
  - Whether the close to verbatim match includes subtle, non-obvious differences.
  - Whether those differences are due to human error or AI error.

| Case Title | Sequence | Judgement Quote | Generative Quote | Comments |
|---|---|---|---|---|
| McLaughlin & Harvey Ltd v Lockton Companies International Ltd | [24] Most disputes which arise relate to the third test and the issue of reasonable sufficiency of notice, namely whether the party tendering the document has done all that was reasonably sufficient to give the other notice of the conditions. Whether they did so or not is a question of fact (Chitty 13-014). It is well established that where the party tendering the document knows it contains an onerous or unusual term it must show that it has been brought fairly and reasonably to the other's attention (Chitty 13-015). As Lord Denning said in J Spurling Ltd v Bradshaw [1956] 1 WLR 461 at 466: | "Some clauses which I have seen ….would need to be printed in red ink on the face of the document with a ~~read~~ hand pointing to it before the notice could be held to be sufficient." | "The more unreasonable the clause, the greater the notice which must be given of it. Some clauses which I have seen would need to be printed in red ink on the face of the document with a red hand pointing to it before the notice could be held to be sufficient." | Close to verbatim with accompanying text.<br><br>However, interestingly the judgement has a typo, which can be validated [11] here, however the model correctly selects red over read, which is an interesting indication where on this occasion, the probabilistic next token selection won over a human spelling mistake.<br><br>Inclusion of "the more unreasonable the clause" on the generative quote isnt ideal, however on this occasion, it likely wouldn't have pulled the advice in another direction or had a material impact. |
| J Spurling Ltd v Bradshaw | Therefore, in finding that the conditions formed part of the contract, it does not seem to me that the learned Judge was out of line with what Mr Justice Bailhache said. Nor that there is anything that controverts what was said by Mr Justice Lush in Crooks v. Allan in 5 Queen's Bench Division, page 38; at page 40, where there is another passage relied upon by Mr Sofer. It seems to me, therefore, that these conditions were a part of the contract. Then in this case the question becomes whether, assuming that the goods were negligently kept by the Plaintiffs, the conditions protect. It seems to me that they plainly do. In Alderslade v. Hendon Laundry, 1945 1 King's Bench, Lord Greene, Master of the Rolls, in his Judgment at page 192, said this: | ~~"It must be remembered that a limitation clause of this kind only applies where the damage, in respect of which the limitation clause is operative, takes place within the four corners of the contract. But there is no room for the application of that principle in the present case, because there is no material for finding that the loss of these handkerchiefs was due to some act by the defendants outside what they had contracted to do"~~ | "In my opinion, the true rule is that the party seeking to rely upon a condition of that kind must show that the condition covers the breach, and that the breach is not due to a failure to take reasonable care, or, in other words, is not due to negligence." | Non-verbatim match with similar semantic intent.<br><br>The generative quote has similar semantic meaning to the judgement quote, including statements like "show the condition covers the breach" and "is not due to negligence". |
| Donoghue v Stevenson | The case of Winterbottom Wright 10 M & W 109 is, on the other hand, an authority that is closely applicable. Owing to negligence in the construction of a carriage it broke down, and a stranger to the manufacture and sale sought to recover damages for injuries which he alleged were due to negligence in the work, and it was held that he had no cause of action either in tort or arising out of contract. This case seems to me to show that the manufacturer of any article is not liable to a third party injured by negligent construction, for there can be nothing in the character of a coach to place it in a special category. It may be noted, also, that in this case Alderson B. said 10 M & W 115 : | "The only safe rule is to confine the right to recover to those who enter into the contract; if we go one step beyond that, there is no reason why we should not go fifty." | "The only safe rule is to confine the right to recover to those who enter into the contract; if we go one step beyond that, there is no reason why we should not go fifty." | Verbatim match with source judgement.<br><br>Perhaps due to the popularity of this case. |
| Gillick v West Norfolk and Wisbech AHA | "I respectfully agree with every word of that and especially with the description of the father's authority as a dwindling right. In J. v. C. [1970] AC 668 Lord Guest and Lord MacDermott referred to the decision in Agar-Ellis, 24 Ch D. 317 as an example of the almost absolute power asserted by the father over his children before the Judicature Act 1873 and plainly thought such an assertion was out of place at the present time: see Lord MacDermott at pp. 703–704. In Reg. v. D. [1984] A.C. 778 Lord Brandon of Oakbrook cited Agar-Ellis as an example of the older view of a father's authority which his Lordship and the other members of the House rejected. In my opinion, the view of absolute paternal authority continuing until a child attains majority which was applied in Agar-Ellis is so out of line with present day views that it should no longer be treated as having any authority. I regard it as a historical curiosity. As Fox L.J. pointed out in the Court of Appeal, ante, p. 141H, the Agar-Ellis cases (1878) 10 Ch D 49; 24 Ch D. 317 seemed to have been regarded as somewhat extreme even in their own day, as they were quickly followed by the Guardianship of Infants Act 1886 | "upon the application of the mother of any infant ~~[whether over 16 or not]~~ make such order as it may think fit regarding the custody of such infant and the right of access thereto of either parent, having regard to the welfare of the infant, and to the conduct of the parents …" (Emphasis added). | "upon the application of the mother of any infant (who may apply without next friend), make such order as it may think fit regarding the custody of such infant and the right of access thereto of either parent, having regard to the welfare of the infant, and to the conduct of the parents, and to the wishes as well of the mother as of the father, and may alter, vary, or discharge such order on the application of either parent, or, after the death of either parent, of any guardian under this Act, and may make such order respecting the costs of the application as it may think just." | Close to verbatim with accompanying text.<br><br>The judge is quoting legislation and then from all the research we have done, looks to add square brackets with his own opinion, inside the quote of the legislation – we cannot find legislation that says, "whether over 16 or not".<br><br>The model correctly includes what the legislation actually says (who may apply without next friend) but the judgement ignored that from the legislation.<br><br>Even between the generative AI output and the only source of data we can find [12] on this legislation there are discrepancies as below.<br><br>"and in every case may make such order respecting the costs of the mother and the liability of the father for the same, or otherwise as to costs, ~~application~~ as it may think just."<br><br>We consider the word "application" rather than "mother and the liability of the father for the same, or otherwise as to costs" as a hallucination, where the model is weighting the previous word "application", and |
| | (49 & 50 Vict.c.27) which, by section 5, provided that the court may:" | | | then "costs of the " … "application" will likely me much more popular as a next token than "mother" which doesn't flow well at all, and is a result of common legalese ellipsis e.g. use of "… for the same" is implicit rather than explicit and is likely to exacerbate the hallucination problem. |
| Case of Steel and Morris v The United Kingdom | 37. Under English law the object of a libel action is to vindicate the plaintiff's reputation and to make reparation for the injury done by the wrongful publication of defamatory statements concerning him or her.<br><br>38. The plaintiff carries the burden of proving "publication". As a matter of law (per Bell J at p. 5 of the judgment in the applicants' case) | ~~"any person who causes or procures or authorises or concurs in or approves the publication of a libel is as liable for its publication as a person who physically hands it or sends it off to another. It is not necessary to have written or printed the defamatory material. All those jointly concerned in the commission of a tort (civil wrong) are jointly and severally liable for it, and this applies to libel as it does to any other tort".~~ | "the publication of a libel takes place when the defamatory matter is comprehended by the reader, listener, or observer. In other words, it is the act of conveying the defamatory meaning to a third person or persons who understand it in that sense." | Non-verbatim match with similar semantic intent.<br><br>The language is different, though the model aims to make the same point, that those guilty is anyone involved in the publication, rather than simply the person who wrote or printed the defamatory material.<br><br>We searched the web for similar output or language from the model but couldn't find anything. In this example, the model understands the follow-on quote, tweaks the synonyms and order of the words, but has similar semantic intent.<br><br>It is also impressive the model was aware enough to output semantic similarity with such a small amount of input text. |
| Jones v Balfour News | What the tribunal's decision boiled down to therefore was this: first, that, as was evident on the face of the documents, Mr Jones did not have two years qualifying service; second, that there was no evidence of any breach of any statutory right that he was asserting or dismissal for asserting such a right and, thirdly, that there was no evidence upon which it could be said that s.100(1)(d) applied. The Chairman was not making a decision as to whether or not she accepted what Mr Jones said about the sun streaming in and making the shop unbearably hot, she was simply deciding that, on the evidence which had been placed before her, there was no case for him to say that he did not require two years service before he was entitled to assert a claim for unfair dismissal and she therefore decided the arguments as to his entitlement against him and it followed that, as had been suggested might happen in the letter of 9th January 1998, he was not someone who could take advantage of the 1996 Act and claim for unfair dismissal.<br><br>It seems to us that those findings were unobjectionable and that no real point of law could be raised | ~~"Where (in accordance with the following provisions of this Part) the Secretary of State makes employment tribunal procedure regulations, the regulations may provide that any act which is required or authorised by the regulations to be done by an employment tribunal and is of a description specified by the regulations for the purposes of this subsection may be done by the person mentioned in subsection (1)(a) alone."~~ | "A tribunal shall be constituted by a chairman sitting alone if it appears to the chairman that the proceedings can be more conveniently or expeditiously disposed of by a chairman sitting alone than by a tribunal constituted in accordance with subsection (1) or (2) and the parties to the proceedings consent." | Non-verbatim match with unrelated intent.<br><br>Model seems completely unaware of the preceding sequence, and it is attempting to compute a follow-on quote by the text we have provided to it.<br><br>This is odd and suggests the model is unaware the case exists. We ask the question of GPT-4 and it replies "*The case of Jones v Balfour News [2000] does not appear to exist in the UK legal system. There may be a similar case with a different name, but without more information, it is not possible to provide the regulations within this specific case.*"<br><br>Is absence of knowledge of a case considered hallucination? In the "provide case law to support my argument" use case, excluding less well-known cases may not cause issue, however "check the other-side's defence for case misrepresentation" might provide odd results if the models are unaware certain case law exists, but can be found on Lexis/Westlaw. |
| | as a result of them. However, it has been said, on behalf of Mr Jones by Ms Stacey, who has represented him extremely competently through the ELAAS scheme, that the decision of the Chairman was a decision which should never have been made because she should have been sitting not as a Chairman alone but with other members of a tribunal.<br><br>It seems to us that that is not a submission that we should accept. A tribunal can be composed of a Chairman sitting alone in certain circumstances. In this instance, it seems to us, that the tribunal could comprise a Chairman sitting alone, because by virtue of s.4(6) of the Employment Tribunals Act 1996 it is provided: | | | Perhaps a legal strategy in future could be to cite cases generative AI is unaware of. |
| AEG (UK) Ltd. v Logic Resource Ltd | In my judgment, the District Judge erred in two respects as a matter of law: first, by failing to apply the Interfoto test as described by Chitty; and, secondly, by treating condition 7.5 in isolation and not in context, and thus adopting a flawed approach to the proper construction of the condition which is also a question of law. I would add that I would also criticise him in a third respect, namely in relation to his reference to insurance, there being no evidence before him that this would have been an insurable risk at the instance of the defendants.<br><br>It follows that the appellants are entitled to succeed in this appeal on the incorporation ground, so that it is not necessary to consider UCTA, save to say that, in my judgment, the respondents, on whom the burden of proof lies under UCTA, must a fortiori fail to satisfy the UCTA reasonableness test. This is because the schedule 2 guidelines, in paragraph (c), require the Court to take into account: | "...whether the customer knew or ought reasonably to have known of the existence and extent of the term (having regard among other things to any custom of the trade and any previous course of dealing between the parties.)" | "the extent to which it was open to the parties to enter into a different contract or to agree to a variation of the term in question, and the extent to which the party subject to the term or the party seeking to rely on it had an opportunity to negotiate the term, taking into consideration the relative bargaining power of the parties and any other relevant circumstances." | Non-verbatim match with similar semantic intent.<br><br>Is "knowing the existence and extent of the term" equivalent to "… had the opportunity to negotiate the term". Is "taking into consideration the relative bargaining power of the parties" equivalent to "having regard among other things to any custom of the trade".<br><br>Knowing and negotiating (if you negotiate, you know) seem closer as synonyms, so the first section is perhaps "semantic enough". "Bargaining power of the parties" doesn't seem to semantically map to "having regard among other things to any custom of the trade" albeit bargaining power could be considered a custom of the trade.<br><br>The model has some semantic understanding of the follow-on quote it seems but is perhaps conflating another similar semantic summary. |
| Barclays Bank Plc v O'Brien | The equitable principle applied in Turnbull v. Duval and Chaplin v. Brammall seems to be that if a wife signs a security document at her debtor husband's request, the creditor will be unable to enforce the security unless either the debtor or the creditor has taken positive steps to try and ensure that the wife understands the import of the security documents or unless she has obtained independent advice. The two cases contrast oddly with Bainbriqqe v. Browne, where no such requirement was imposed on the father's mortgagees and, since there was no undue influence or misrepresentation by the debtor, cannot, in my opinion, be explained by treating the debtor as the creditor's agent.<br><br>In Howes v. Bishop [1909] 2 KB 390 the creditor obtained iudgment against a debtor and it was then agreed that the debtor and his wife would give a joint and several promissory note for the amount of the judgment. The wife signed the note. The jury found that she understood what she was doing but that her signature had been procured by her husband's influence. The jury did not, however, find undue influence. The wife had no | ~~"I must not be understood to be laying down a rule that in no case where a wife acts on her husband's instructions and under his influence is it necessary to show that she has received independent advice ... but, in my opinion, there cannot be a rule of universal application that the rules of the Courts of Equity as to confidential relationships necessarily apply to the relation of husband and wife".~~ | "It is quite true that if a person in a fiduciary position obtains a security from a person under his influence, the security is not binding unless the person giving it has had independent advice. But that is not this case. Here the wife was not giving a security; she was giving a joint and several promissory note. The jury have found that she understood what she was doing, and that there was no undue influence. In these circumstances, I think that the wife is bound by her signature." | Non-verbatim match with similar semantic intent.<br><br>Although there is similar intent, it looks like the model is treating the input as a question, ignoring the instruction to continue the sequence as it has done before. Searching "she was giving a joint and several promissory note." does bring up Barclays v O'Brien on Bailli.<br><br>For some reason, the model has missed the "intent" of the prompt on this occasion. |
| | independent advice. The Court of Appeal held that the wife was bound.<br><br>Lord Alverstone C.J. said at page 395: | | | |
| Butler Machine Tool Co Ltd. v Ex-Cell-O Corp (England) Ltd. | If those documents are analysed in our traditional method, the result would seem to me to be this: The quotation of the 23rd May, 1969 was an offer by the sellers to the buyers containing the terms and conditions on the back. The order of the 27th May, 1969 purported to be an acceptance of that offer in that it was for the same machine at the same price, but it contained such additions as to cost of installation, date of delivery and so forth that it was in law a rejection of the offer and constituted a counter-offer. As Mr. Justice Megaw said in Trollope & Colls Ltd. & ors. v. Atomic Power Constructions Ltd. [1963] 1 W.L.R. 333 at page 337: | ~~"... the counter-offer kills the original offer".~~ | "An acceptance which is not in the terms of the offer is not an acceptance at all; it is a counter-offer, which must be accepted by the original offeror before a contract is made." | Non-verbatim match with similar semantic intent.<br><br>What is quite fascinating about this example, is no where online can we find sections of the generative quote. The model is clearly reasoning across all summaries and the verbatim judgement that is referenced (provided below) to articulate the same meaning, but in a more verbose way. Perhaps due to model temperature with the judgement quote being too prescriptive.<br><br>The […] from the judgement "To make a contract, there must be an offer and an acceptance. If an offer is rejected, there may be a counter-offer, but **the counter-offer kills the original offer.**"<br><br>Swapping out the generative quote with the verbatim text from the judgement shouldn't impact the overall meaning of the judgement referencing this point. |
| Farley v Skinner | 24. Interpreting the dicta of Bingham LJ in Watts v Morrow narrowly the Court of Appeal in Knott v Bolton ruled that the central object of the contract was to design a house, not to provide pleasure to the occupiers of the house. It is important, however, to note that Knott v Bolton was decided a few months before the decision of the House in Ruxley Electronics and Construction Ltd v Forsyth [1996] AC 344. In any event, the technicality of the reasoning in Knott v Bolton, and therefore in the Court of Appeal judgments in the present case, is apparent. It is obvious, and conceded, that if an architect is employed only to design a staircase, or a surveyor is employed only to investigate aircraft noise, the breach of such a distinct obligation may result in an award of non-pecuniary damages. Logically the same must be the case if the architect or surveyor, apart from entering into a general retainer, concludes a separate contract, separately remunerated, in respect of the design of a staircase or the investigation of aircraft noise. If this is so the distinction drawn in Knott v Bolton and in the present case is a matter of form and not substance. David Capper, | "A ruling that intangible interests only qualify for legal protection where they are the "very object of the contract" is tantamount to a ruling that contracts where these interests are merely important, but not the central object of the contract, are in part unenforceable. It is very difficult to see what policy objection there can be to parties to a contract agreeing that these interests are to be protected via contracts where the central object is something else. If the defendant is unwilling to accept this responsibility he or she can say so and either no contract will be made or one will be made but including a disclaimer." | "It is submitted that the distinction between contracts whose central object is to provide pleasure, relaxation or peace of mind and those which are not is unprincipled and unworkable. The better view is that, in principle, damages for non-pecuniary loss should be recoverable for breach of any contract if it was reasonably foreseeable that such loss would be suffered as a result of the breach." | Non-verbatim match with unrelated intent<br><br>The generative quote seems to be conflating [13] sections of other quotes within the judgements e.g. "object of a contract is to provide pleasure, relaxation, peace of mind …" |
| | "Damages for Distress and Disappointment - The Limits of Watts v Morrow" (2000) 116 LQR 553, 556) has persuasively argued: | | | |
| Prest v Petrodel Resources Limited and others | "8. Subject to very limited exceptions, most of which are statutory, a company is a legal entity distinct from its shareholders. It has rights and liabilities of its own which are distinct from those of its shareholders. Its property is its own, and not that of its shareholders. In Salomon v A Salomon and Co Ltd [1897] AC 22, the House of Lords held that these principles applied as much to a company that was wholly owned and controlled by one man as to any other company. In Macaura v Northern Assurance Co Ltd [1925] AC 619, the House of Lords held that the sole owner and controller of a company did not even have an insurable interest in property of the company, although economically he was liable to suffer by its destruction. Lord Buckmaster, at pp 626–627 said:" | ~~“no shareholder has any right to any item of property owned by the company, for he has no legal or equitable interest therein. He is entitled to a share in the profits while the company continues to carry on business and a share in the distribution of the surplus assets when the company is wound up.”~~ | 'The company is at law a different person altogether from the subscribers to the memorandum; and, though it may be that after incorporation the business is precisely the same as it was before, and the same persons are managers, and the same hands receive the profits, the company is not in law the agent of the subscribers or trustee for them. Nor are the subscribers as members liable, in any shape or form, except to the extent and in the manner provided by the Act.'" | Non-verbatim match with similar semantic intent.<br><br>This is very interesting. The model has pulled "verbatim" a quote from Salomon v A Salomon which has in effect similar semantic intent.<br><br>In Salomon, the preceding quote says<br>"*I cannot understand how a body corporate thus made “capable” by statute can lose its individuality by issuing the bulk of its capital to one person*" which binds semantically to "*wholly owned and controlled by one man as any other company*".<br><br>The model is wrong based on the instruction, but the hallucination is sensible. |
| Blackpool and Fylde Aero Club Limited v Blackpool Borough Council | I readily accept that contracts are not to be lightly implied. Having examined what the parties said and did, the court must be able to conclude with confidence both that the parties intended to create contractual relations and that the agreement was to the effect contended for. It must also, in most cases, be able to answer the question posed by Mustill LJ in The Kapetan Markos N.L. (NO.2) [1987] 2 Ll. 321 at 331: | ~~"What was the mechanism for offer and acceptance?"~~ | "What is the contract? What are its terms? What are the parties' respective obligations? What are the consequences of a failure to perform?" | Non-verbatim match with unrelated intent<br><br>It seems the model is weighting the introduction of "*be able to answer the question posed by …*" over the context of the legislation. This is perhaps resulting in shotgunning of multiple questions. |
| Secretary of State for Health and another v Servier Laboratories Ltd and others | 34. As to the tort's essential elements, Lord Hoffmann addressed first what constitutes unlawful means. It was primarily on this issue that Lord Nicholls disagreed with the majority.<br>35. Lord Hoffmann began his analysis as follows:<br>"45. The most important question concerning this tort is what should count as unlawful means. It will be recalled that in Allen v Flood [1898] AC 1, 96 , Lord Watson described the tort thus: | 'when the act induced is within the right of the immediate actor, and is therefore not wrongful in so far as he is concerned, it may yet be to the detriment of a third party; and in that case …the inducer may be held liable if he can be shewn to have procured his object by the use of illegal means directed against that third party.' | "46. 'The wrongful acts which are the subject of this action consist in maliciously inducing a person to break a contract with another, or to abstain from entering into a contract with another, or to refuse to employ another, or to dismiss him from his employment, or to do any other act whereby another may sustain damage, by means of threats, intimidation, or molestation, or by other unlawful means.'" | Non-verbatim match with unrelated intent<br><br>Although there are some similar words like "induce" and "act", the deviation for the text it too significant that we don't believe it would be appropriate to say they are semantically similar, however, in defining "unlawful means" both could be considered definitions of that term, albeit not similar definitions. |
| London Borough of Merton Council v Nuffield Health | Although the ISC case has not been without its academic critics: see Tudor on Charities, 11th ed (2023), para 1-182, this aspect of the Upper Tribunal's reasoning is, as the editors of Tudor note, well supported by earlier authority. In Jones v Williams (1767) 2 Amb 651 the gift was for supplying the inhabitants of Chepstow (not just the poor inhabitants) with water. Holding that purpose to be charitable, Lord Hardwicke LC said (at p 652): | ~~"definition of charity; a gift to a general public use, which extends to the poor as well as to the rich."~~ | "Nothing can be more beneficial to society than the improvement of the health and convenience of the inhabitants; and if the purpose is beneficial to the public, it is a charity." | Non-verbatim match with unrelated intent<br><br>There is similarity, however the key point of the judgement text is that charity extends too rich and poor. Like the previous example, both could pass for describing "*the purpose to be charitable*". |
| The University of Dundee v Prasun Chakraborty | 10. The protection of confidential communications between client and lawyer lay at the heart of legal professional privilege ( Ventouris v Mountain [1991] 1 WLR 607 at 475 ). Communications between clients and lawyers should be "secure against the possibility of any scrutiny from others" ( Three Rivers DC v Bank of England (No. 6) [2005] 1 AC 610 at para 34 ). There had to be a "relevant legal context" in which the communications were made ( ibid at paras 38, 62 and 111). The advice could relate to the rights, liabilities, obligations or remedies of the client under private or public law ( ibid ). In this case the advice related to the interpretation and sufficiency of the matters discussed in the original report. The advice may relate partly to the rights or obligations of others ( ibid at para 56). It may not be strictly legal but could be presentational or revisal. | 10. The protection of confidential communications between client and lawyer lay at the heart of legal professional privilege ( Ventouris v Mountain [1991] 1 WLR 607 at 475 ). Communications between clients and lawyers should be "secure against the possibility of any scrutiny from others" ( Three Rivers DC v Bank of England (No. 6) [2005] 1 AC 610 at para 34 ). There had to be a "relevant legal context" in which the communications were made ( ibid at paras 38, 62 and 111). The advice could relate to the rights, liabilities, obligations or remedies of the client under private or public law ( ibid ). In this case the advice related to the interpretation and sufficiency of the matters discussed in the original report. The advice may relate partly to the rights or obligations of others ( ibid at para 56). It may not be strictly legal but could be presentational or revisal. | 10. The protection of confidential communications between client and lawyer lay at the heart of legal professional privilege ( Ventouris v Mountain [1991] 1 WLR 607 at 475 ). Communications between clients and lawyers should be "secure against the possibility of any scrutiny from others" ( Three Rivers DC v Bank of England (No. 6) [2005] 1 AC 610 at para 34 ). There had to be a "reasonable expectation of confidentiality" in which the communications were made ( ibid at paras 38, 62 and 111). The advice could relate to the rights, liabilities, obligations or remedies of the client under private or public law ( ibid ). In this case the advice related to the interpretation and sufficiency of the matters discussed in the original report. The advice may relate partly to the rights or obligations of others ( ibid at para 56). It may not be strictly legal but could be presentational or revisal. | Non-verbatim match with unrelated intent<br><br>Using a slightly different component of OpenAI (Insert) we now begin testing the inserting of blanks to detect whether the model is fully aware of the context. This uses text-davinci-003 which is less sophisticated than GPT-4. GPT-4 is unavailable in Insert within Playground.<br><br>We can see this highlights the subtle, non-obvious hallucination problem, which wont always be obvious. Within the context of the judgement quote, the generative replacement of "reasonable expectation of confidentiality" arguably makes more sense than "relevant legal context". |
| Pizza Express Group Ltd v Liberty Mutual Insurance Europe SE | 27. There was no dispute as to the applicable principles of construction. They have been addressed in a number of the COVID-19 BI insurance authorities by reference to previous Supreme Court authority, including Wood v Capita Insurance Services Ltd [2017] AC 1173 and Arnold v Britton [2015] AC 1619 . The essential principles are as follows:<br>i) The Policy must be construed objectively by asking what a reasonable policyholder, with all the background knowledge which would reasonably have been available to both parties when they entered into the contract, would have understood the language of the Policy to mean.<br>ii) This does not involve "a literalist exercise focussed solely on a parsing of the wording of the particular clause": Wood v Capita at [10]. Instead, it is essential to construe contractual words in their applicable context. Their meaning must be assessed in the context of the clause in which they appear as well as in the landscape of the document as a whole.<br>iii) The unitary exercise of contractual construction can require the court to give weight to the implications of rival | 27. There was no dispute as to the applicable principles of construction. They have been addressed in a number of the COVID-19 BI insurance authorities by reference to previous Supreme Court authority, including Wood v Capita Insurance Services Ltd [2017] AC 1173 and Arnold v Britton [2015] AC 1619 . The essential principles are as follows:<br>i) The Policy must be construed objectively by asking what a reasonable policyholder, with all the background knowledge which would reasonably have been available to both parties when they entered into the contract, would have understood the language of the Policy to mean.<br>ii) This does not involve "a literalist exercise focussed solely on a parsing of the wording of the particular clause": Wood v Capita at [10]. Instead, it is essential to construe contractual words in their applicable context. Their meaning must be assessed in the context of the clause in which they appear as well as in the landscape of the document as a whole.<br>iii) The unitary exercise of contractual construction can require the court to give weight to the implications of rival | 27. There was no dispute as to the applicable principles of construction. They have been addressed in a number of the COVID-19 BI insurance authorities by reference to previous Supreme Court authority, including Wood v Capita Insurance Services Ltd [2017] AC 1173 and Arnold v Britton [2015] AC 1619 . The essential principles are as follows:<br>i) The Policy must be construed objectively by asking what a reasonable policyholder, with all the background knowledge which would reasonably have been available to both parties when they entered into the contract, would have understood the language of the Policy to mean.<br>ii) This does not involve "a search for the intention of the parties": Wood v Capita at [10]. Instead, it is essential to construe contractual words in their applicable context. Their meaning must be assessed in the context of the clause in which they appear as well as in the landscape of the document as a whole.<br>iii) The unitary exercise of contractual construction can require the court to give weight to the implications of rival constructions by reaching a view as to which construction is more | Non-verbatim match with similar semantic intent.<br><br>This is semantically similar in that it is an opposite. This does not involve "*a literalist exercise focussed solely on a parsing of the wording of the particular clause*" it to an extent the opposite of "*a search for the intention of the parties*" e.g., the analysis should not be literal focussed only on clause wording, but "instead" it should focus on words within context … searching for the intention of the parties.<br><br>"Searching for the intention of the parties" is very common run of text, and it is relevant to the entire section of text from the judgement, although different to the verbatim judgement. |
| | constructions by reaching a view as to which construction is more consistent with commercial common sense. However, commercial common sense should not be invoked retrospectively, or to rewrite a contract, in an attempt to assist an unwise party or to penalise an astute party. | constructions by reaching a view as to which construction is more consistent with commercial common sense. However, commercial common sense should not be invoked retrospectively, or to rewrite a contract, in an attempt to assist an unwise party or to penalise an astute party. | consistent with commercial common sense. However, commercial common sense should not be invoked retrospectively, or to rewrite a contract, in an attempt to assist an unwise party or to penalise an astute party. | |
| JTI Polska Sp Zoo v Jakubowski | 15. The trial of the respondents' claim for the Excise Duty was heard by Judge Pelling KC ("the judge"), sitting as a High Court Judge, on 26 May 2021. The appellants accepted that in light of the Buchanan decision the judge was bound to hold that the Excise Duty was recoverable under article 23.4 of the CMR but they contended that the decision was wrong and should be departed from. They accordingly made an application for a certificate under section 12 of the Administration of Justice Act 1969 that the case was suitable for an appeal directly to the Supreme Court. The judge granted the certificate, principally on the basis of criticism of Buchanan by the leading English commentators on the CMR and the uncertainty created by the decision of the Court of Appeal in Sandeman Coprimar SA v Transitos y Transportes Integrales SL [2003] EWCA Civ 113; [2003] QB 1270 (Sandeman). In that decision Buchanan was criticised and distinguished and it was stated that the decision should not be "applied any more widely by the courts of this country than respect for the doctrine of precedent requires" (para 38). | 15. The trial of the respondents' claim for the Excise Duty was heard by Judge Pelling KC ("the judge"), sitting as a High Court Judge, on 26 May 2021. The appellants accepted that in light of the Buchanan decision the judge was bound to hold that the Excise Duty was recoverable under article 23.4 of the CMR but they contended that the decision was wrong and should be departed from. They accordingly made an application for a certificate under section 12 of the Administration of Justice Act 1969 that the case was suitable for an appeal directly to the Supreme Court. The judge granted the certificate, principally on the basis of criticism of Buchanan by the leading English commentators on the CMR and the uncertainty created by the decision of the Court of Appeal in Sandeman Coprimar SA v Transitos y Transportes Integrales SL [2003] EWCA Civ 113; [2003] QB 1270 (Sandeman). In that decision Buchanan was criticised and distinguished and it was stated that the decision should not be "applied any more widely by the courts of this country than respect for the doctrine of precedent requires" (para 38). | 15. The trial of the respondents' claim for the Excise Duty was heard by Judge Pelling KC ("the judge"), sitting as a High Court Judge, on 26 May 2021. The appellants accepted that in light of the Buchanan decision the judge was bound to hold that the Excise Duty was recoverable under article 23.4 of the CMR but they contended that the decision was wrong and should be departed from. They accordingly made an application for a certificate under section 12 of the Administration of Justice Act 1969 that the case was suitable for an appeal directly to the Supreme Court. The judge granted the certificate, principally on the basis of criticism of Buchanan by the leading English commentators on the CMR and the uncertainty created by the decision of the Court of Appeal in Sandeman Coprimar SA v Transitos y Transportes Integrales SL [2003] EWCA Civ 113; [2003] QB 1270 (Sandeman). In that decision Buchanan was criticised and distinguished and it was stated that the decision should not be "followed without careful consideration" (para 38). | Non-verbatim match with similar semantic intent.<br><br>"*Followed without careful consideration*" is not wrong, as a broad high-level explanation for additional "*charges*" that can be recovered from a carrier under Article 23.4 of the CMR, with different EU jurisdictions accepting either a wide or narrow definition of "*charges*". Therefore, "*careful consideration*" is valid, albeit far from the more specific "*applied any more widely*" statement, which the courts in England detailed as a narrower definition.<br><br>Note: JTI was heard after the 2021 cut-off date for the GPT models, which could play a role here, though Buchanan is well documented and was heard over 20 years ago. |
| DSG Retail Limited v Mastercard Incorporated | 87. The next step is to consider the applicable English law as to the nature of a cause of action.<br>88. In Paragon Finance (1998), Millett LJ recalled the classic definitions of a cause of action as follows at page 405:<br>"The classic definition of a cause of action was given by Brett J in Cooke v Gill (1873) LR 8 CP 107 [" Cooke v. Gill "] at p. 116:-<br>"Cause of action" has been held from the earliest time to mean every fact which is material to be proved to entitle the plaintiff to succeed - every fact which the defendant would have a right to traverse" (my emphasis). | 87. The next step is to consider the applicable English law as to the nature of a cause of action.<br>88. In Paragon Finance (1998), Millett LJ recalled the classic definitions of a cause of action as follows at page 405:<br>"The classic definition of a cause of action was given by Brett J in Cooke v Gill (1873) LR 8 CP 107 [" Cooke v. Gill "] at p. 116:-<br>"Cause of action" has been held from the earliest time to mean every fact which is material to be proved to entitle the plaintiff to succeed - every fact which the defendant would have a right to traverse" (my emphasis). | 87. The next step is to consider the applicable English law as to the nature of a cause of action.<br>88. In Paragon Finance (1998), Millett LJ recalled the classic definitions of a cause of action as follows at page 405:<br>"The classic definition of a cause of action was given by Brett J in Cooke v Gill (1873) LR 8 CP 107 [" Cooke v. Gill "] at p. 116:-<br>"A cause of action is a factual situation the existence of which entitles one person to obtain from the court a remedy against another person." (my emphasis). | Non-verbatim match with similar semantic intent.<br><br>The generative "cause of action" quote is a valid definition, however comes from another judgement, [14] [Letang v Cooper] with one subtle difference, the exclusion of the word "simply":<br><br>"A cause of action is ~~simply~~ a factual situation the existence of which entitles one person to obtain from the court a remedy against another person."<br><br>Interestingly, research on Google shows both versions are used, with the version that includes "simply" much more popular. This suggests the model has decided, as some humans have done who are repeating this judgement language, that "simply" is not material and can be safely excluded for clarity. |
| Harlow Higinbotham (formerly BWK) v Teekhungam & another | 46. The Court has jurisdiction to stay or strike out a claim where no real or substantial wrong has been committed and litigating the claim will yield no tangible or legitimate benefit to the claimant proportionate to the likely costs and use of court procedures: in other words, " the game is not worth the candle ": Jameel [69]-[70] per Lord Phillips MR; Schellenberg -v- BBC [2000] EMLR 296 , 319 per Eady J. The jurisdiction is useful where a claim " is obviously pointless or wasteful ": Vidal-Hall -v- Google Inc [2016] QB 1003 [136] per Lord Dyson MR. Although Jameel was a defamation claim (and defamation claims present particular features) the jurisdiction is of general application: Sullivan -v- Bristol Film Studios Limited [2012] EMLR 27 and has been held to extend to malicious falsehood claims: Niche Products [63] and c.f. Tesla Motors [47]-[49]. | 46. The Court has jurisdiction to stay or strike out a claim where no real or substantial wrong has been committed and litigating the claim will yield no tangible or legitimate benefit to the claimant proportionate to the likely costs and use of court procedures: in other words, " the game is not worth the candle ": Jameel [69]-[70] per Lord Phillips MR; Schellenberg -v- BBC [2000] EMLR 296 , 319 per Eady J. The jurisdiction is useful where a claim " is obviously pointless or wasteful ": Vidal-Hall -v- Google Inc [2016] QB 1003 [136] per Lord Dyson MR. Although Jameel was a defamation claim (and defamation claims present particular features) the jurisdiction is of general application: Sullivan -v- Bristol Film Studios Limited [2012] EMLR 27 and has been held to extend to malicious falsehood claims: Niche Products [63] and c.f. Tesla Motors [47]-[49]. | 46. The Court has jurisdiction to stay or strike out a claim where no real or substantial wrong has been committed and litigating the claim will yield no tangible or legitimate benefit to the claimant proportionate to the likely costs and use of court procedures: in other words, "vexatious litigation": Jameel [69]-[70] per Lord Phillips MR; Schellenberg -v- BBC [2000] EMLR 296 , 319 per Eady J. The jurisdiction is useful where a claim " is obviously pointless or wasteful ": Vidal-Hall -v- Google Inc [2016] QB 1003 [136] per Lord Dyson MR. Although Jameel was a defamation claim (and defamation claims present particular features) the jurisdiction is of general application: Sullivan -v- Bristol Film Studios Limited [2012] EMLR 27 and has been held to extend to malicious falsehood claims: Niche Products [63] and c.f. Tesla Motors [47]-[49]. | Non-verbatim match with similar semantic intent.<br><br>The term "vexatious" defines the text that precedes the quote:<br><br>"*denoting an action or the bringer of an action that is brought without sufficient grounds for winning, purely to cause annoyance to the defendant.*"<br><br>However, this does not refer correctly to the quote from the judgement. |
| Williamson v Bishop of London | 31. Mr Wynne pointed out correctly that section 42 does not specify the consequences of a claim brought in breach of a CPO. Moreover, although on the face of it section 42 appears to envisage that leave will be a condition precedent to the institution of proceedings (save in relation to existing proceedings at the time of the CPO), he pointed out that this is achieved by saying that that will be the nature of the order made by way of a CPO, rather than by providing for it directly (unlike section 139(2) MHA ). Accordingly, and since court orders with conditions precedent (like unless orders) have always been subject to relief from sanction, he submitted that the same must be true here. These are legitimate points to make, but it does not follow that that was the statutory intention. No doubt section 42 had to take this form because CPOs are to be granted on a litigant by litigant basis. In any event, I do not consider that this distinction can dictate the right answer. Nor is the statutory language on its own determinative, as Lord Bingham in Seal made clear.<br>32. As Lord Bingham explained in R (Quintavalle) v Secretary of | 31. Mr Wynne pointed out correctly that section 42 does not specify the consequences of a claim brought in breach of a CPO. Moreover, although on the face of it section 42 appears to envisage that leave will be a condition precedent to the institution of proceedings (save in relation to existing proceedings at the time of the CPO), he pointed out that this is achieved by saying that that will be the nature of the order made by way of a CPO, rather than by providing for it directly (unlike section 139(2) MHA ). Accordingly, and since court orders with conditions precedent (like unless orders) have always been subject to relief from sanction, he submitted that the same must be true here. These are legitimate points to make, but it does not follow that that was the statutory intention. No doubt section 42 had to take this form because CPOs are to be granted on a litigant by litigant basis. In any event, I do not consider that this distinction can dictate the right answer. Nor is the statutory language on its own determinative, as Lord Bingham in Seal made clear.<br>32. As Lord Bingham explained in R (Quintavalle) v Secretary of | 31. Mr Wynne pointed out correctly that section 42 does not specify the consequences of a claim brought in breach of a CPO. Moreover, although on the face of it section 42 appears to envisage that leave will be a condition precedent to the institution of proceedings (save in relation to existing proceedings at the time of the CPO), he pointed out that this is achieved by saying that that will be the nature of the order made by way of a CPO, rather than by providing for it directly (unlike section 139(2) MHA ). Accordingly, and since court orders with conditions precedent (like unless orders) have always been subject to relief from sanction, he submitted that the same must be true here. These are legitimate points to make, but it does not follow that that was the statutory intention. No doubt section 42 had to take this form because CPOs are to be granted on a litigant by litigant basis. In any event, I do not consider that this distinction can dictate the right answer. Nor is the statutory language on its own determinative, as Lord Bingham in Seal made clear.<br>32. As Lord Bingham explained in R (Quintavalle) v Secretary of | Non-verbatim match with similar semantic intent.<br><br>The quote "mischief at which the statute was aimed" seems to be odd language however is commonly used to describe the purpose of a statute: used in some other judgements [15] here and [16] here.<br><br>The summary does a reasonable job of representing the quote from the judgement. |
| | State for Health [2003] UKHL 13; [2003] 2 AC 687 at paragraph 8 : "Every statute other than a pure consolidating statute is, after all, enacted to make some change, or address some problem, or remove some blemish, or effect some improvement in the national life. The court's task, within the permissible bounds of interpretation, is to give effect to Parliament's purpose. So the controversial provisions should be read in the context of the statute as a whole, and the statute as a whole should be read in the historical context of the situation which led to its enactment." | State for Health [2003] UKHL 13; [2003] 2 AC 687 at paragraph 8 : "Every statute other than a pure consolidating statute is, after all, enacted to make some change, or address some problem, or remove some blemish, or effect some improvement in the national life. The court's task, within the permissible bounds of interpretation, is to give effect to Parliament's purpose. So the controversial provisions should be read in the context of the statute as a whole, and the statute as a whole should be read in the historical context of the situation which led to its enactment." | State for Health [2003] UKHL 13; [2003] 2 AC 687 at paragraph 8 : "The court must look at the language used, the context in which it is used, the purpose of the statute and the material available to show the mischief at which the statute was aimed." | |

## 3.1 Discussion

The results are broken down as follows:

| Verbatim match with the source judgement. | Close to verbatim match including accompanying text within the source judgement. | Non-verbatim match with similar semantic intent. | Non-verbatim match with unrelated intent. |
|---|---|---|---|
| 1 | 2 | 11 | 6 |

Based on twenty trials, we found more than fifty percent of the results that were non-verbatim matches with similar semantic intent. These outcomes indicate that the statements replaced by the model retained the same or similar meaning, though in the majority of cases, they were concise, summarized variations of the original language, as opposed to verbatim quotes.

Our experimentation with the model revealed a solitary, flawless match in the context of Donoghue v Stevenson. We attribute this success, in part, to the case's widespread notoriety. The quote in question - "*The only safe rule is to confine the right to recover to those who enter into the contract; if we go one step beyond that, there is no reason why we should not go fifty*" - is routinely cited online within a variety of contexts, including [17] case summaries, [18] judicial opinions and also [19] academic publications. This data suggests that the model places greater emphasis on well-known quotes.

We experienced two closely matched instances:

In McLaughlin v Lockton, we encountered an anecdotal example where the probabilistic model correctly guessed the next word, surpassing a human made error within the judgement. However, such cases are exceptions rather than the rule.

In Gillick v West Norfolk presented a more complex scenario where the court's commentary within square brackets resulted in contaminating the quoted legislation. The model, albeit ignoring the judge's remarks, replaced the square brackets with a component excluded from the judgment but found within the legislation. While the model's rendition of the legislative preamble was more accurate, the subsequent text provided by the model exhibited factual inaccuracies in comparison to the statute. This exemplifies the nuanced and complicated nature of the hallucination problem in legal research.

It is worth mentioning that the majority of our evaluations produced non-verbatim matches, either with comparable semantic intent - signifying the same or nearly the same meaning - or with completely unrelated intent, a clear indication of the model's hallucinative tendencies.

It's our argument that:

**Verbatim match with source judgement** – Achieving a verbatim match with the source judgment can be considered fortunate. However, the degree of success often depends on various factors, such as the extent to which the core judgment is condensed in online summaries. Consequently, securing an exact match is something of a game of chance, especially in the case of popular instances where the underlying case law is clear and straightforward, allowing for concise summaries that do not stray from the verbatim language of the judgment.

**Close to verbatim match including accompanying text –** Obtaining a close-to-verbatim match, including the accompanying text, can occur in several forms. In the context of McLaughlin, the error was on the court's side, indicating that the model is not always responsible for the hallucination of data - although such outcomes should not be relied upon consistently. Alternatively, Gillick highlights how courts sometimes modify quotes from previous judgments or legislation, demonstrating how even slight contamination of data could be a significant challenge for LLMs seeking to understand and interpret information accurately. Additionally, this also emphasizes the need for models to comprehend square brackets, a crucial aspect in the legal industry for maintaining accuracy in the data they process.

**Non-verbatim match with similar semantic intent –** One of the most interesting yet predictable challenges associated with training LLMs on data from the entire internet is achieving non-verbatim matches while retaining similar semantic intent. This issue is rather significant since it necessitates an understanding of case law, legislation, and summaries of that data. In computer science parlance, we can draw a comparison between common law and machine code and case summaries and higher-level languages. We speculate that if models were trained with machine code and equivalent higher-level language, hallucination and "code contamination" would arise on a considerable scale. It's possible that the law may progress to a point where only court-sourced legal data is utilised, given that summary and interpretation are unnecessary, and case law and legislation explicitly satisfy the terms of deterministic code.

**Non-verbatim match with unrelated intent** – Non-verbatim matching with unrelated intent represents the endpoint of the hallucination spectrum. Nonetheless, it presents the greatest potential for subtlety and ambiguity, contingent on the level of output from the model. For instance, in Dundee v Prasan, the term "relevant legal context" could be substituted for "reasonable expectation of confidentiality" without raising any alarms, particularly in the case of confidential communications. Such a shift may go unnoticed, emphasising how careful evaluation is essential even for seemingly minor changes to determine whether the intended meaning remains intact.

To summarise, our paper highlights that hallucination of data in legal language is a nuanced problem and should not be dismissed as an over-arching fictitious case law issue. Our experiments reveal that neither consumer nor enterprise-level use of ChatGPT delivers the precise controls necessary for mitigating exposure effectively. Even the most careful and astute lawyers may fall prey to some of these hallucinatory outcomes, underlining the need for the legal industry to consider the acquisition and application of sufficient safeguards.

We would like to preface that we are not against the application of LLMs. More specifically, regarding the Reasoning side of the model, we are enthusiastic about the opportunities it presents. Our experiments indicate that it performs quite effectively in volume-type assignments such as verification, contract review, and dispute resolution. We believe that advancements in LLM research in this area should be brought to market as quickly as possible. However, if we disregard the issue of common law contamination and the need to limit the use of LLMs for legal research, their use in the industry could rapidly be rendered unlawful, depriving us of a significant AI-driven advancement in legal reasoning.

Ultimately, we believe that entities such as Lexis, Thomson Reuters, and Casetext possess a significant advantage as the primary custodians of the most critical legal data, i.e., common law, legislative and regulatory rules. Although our team would like to enhance the capabilities of YCNBot to provide "hallucination verification" research, we do not have access to "justice data." However, our AI team has shared their findings with the [20] Open Justice Data movement. We encourage others to do the same, working together to support open-source practitioners and help improve these technologies for the betterment of all.

## 5 YCNBot Enhancement

Part of our AI strategy is committed to open sourcing functionality within YCNBot which supports the ongoing experimentation and education of LLMs in a safe and balanced way. Due to some of the subtle, not obvious errors we have discussed within this paper, we have released a new feature to YCNBot which has now been open sourced, to detect case law and where it exists in output, block the "Copy and Paste" functionality, working to mitigate any risk. The check is based on regular expressions and common case name formats, as we lack direct access to a case law database. This is also on top of guidance and training videos we do internally, making it clear the system should not be used for Legal Research. We will be looking to add a similar "legislation detection" algorithm which will have the same effect.

You can now clone the most recent version of YCNBot with this feature here:

https://github.com/Travers-Smith/YCNBot

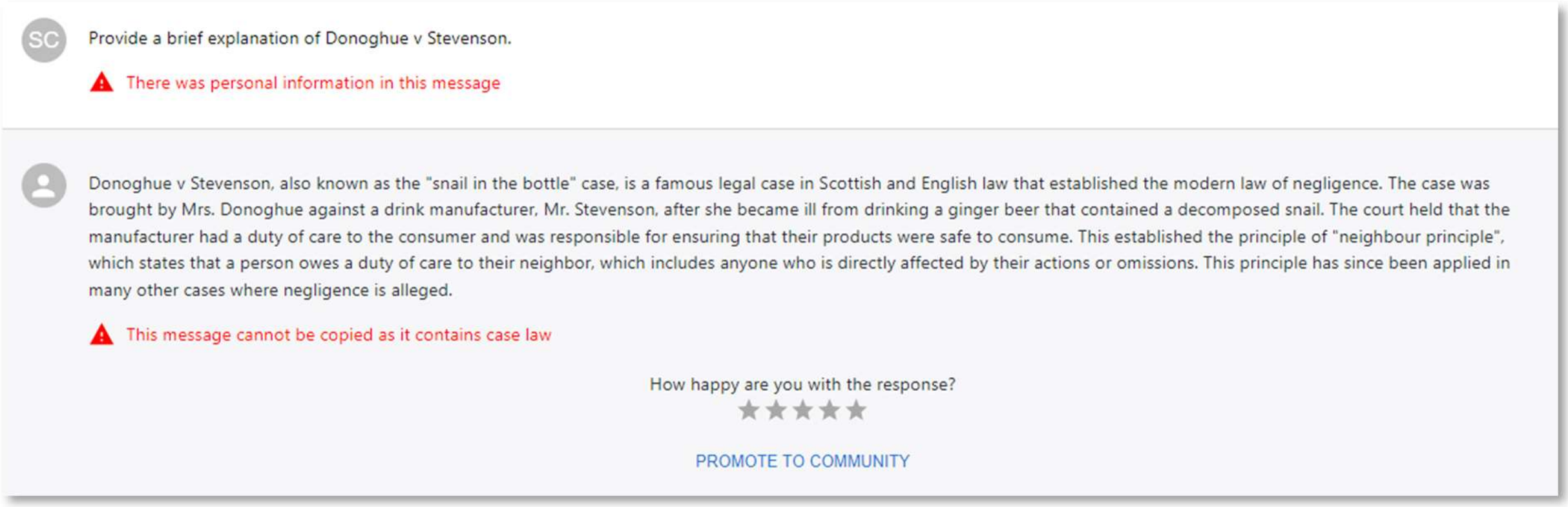